\documentclass[sigconf]{acmart}

\usepackage{booktabs} 
\usepackage{algorithm2e}
\usepackage{algorithmic}

\copyrightyear{2019}
\acmYear{2019} 
\setcopyright{iw3c2w3}
\acmConference[WWW '19 Companion]{Companion Proceedings of the 2019 World Wide Web Conference}{May 13--17, 2019}{San Francisco, CA, USA}
\acmBooktitle{Companion Proceedings of the 2019 World Wide Web Conference (WWW '19 Companion), May 13--17, 2019, San Francisco, CA, USA}
\acmPrice{}
\acmDOI{10.1145/3308560.3316584}
\acmISBN{978-1-4503-6675-5/19/05}

\usepackage{balance}

\begin{document}

\title{Contextual Compositionality Detection with External Knowledge Bases and Word Embeddings}

\author{Dongsheng Wang}
\affiliation{%
  \institution{Department of Computer Science\\University of Copenhagen}
  \city{Copenhagen}
  \country{Denmark}
}
\email{wang@di.ku.dk}

\author{Qiuchi Li}
\affiliation{%
  \institution{Department of Information Engineering\\University of Padua}
  \city{Padova}
  \country{Italy}
}
\email{qiuchili@dei.unipd.it}

\author{Lucas Chaves Lima}
\affiliation{%
  \institution{Department of Computer Science\\University of Copenhagen}
  \city{Copenhagen}
  \country{Denmark}}
\email{lcl@di.ku.dk}

\author{Jakob Grue Simonsen}
\affiliation{%
  \institution{Department of Computer Science\\University of Copenhagen}
  \city{Copenhagen}
  \country{Denmark}
}
\email{simonsen@di.ku.dk}
\author{Christina Lioma}
\affiliation{%
 \institution{Department of Computer Science\\University of Copenhagen}
 \city{Copenhagen}
 \country{Denmark}
}
\email{c.lioma@di.ku.dk}

\renewcommand{\shortauthors}{D. Wang et al.}

\begin{abstract}
When the meaning of a phrase cannot be inferred from the individual meanings of its words (e.g., \textit{hot dog}), that phrase is said to be \textit{non-compositional}. Automatic compositionality detection in multi-word phrases is critical in any application of semantic processing, such as search engines \cite{LiomaLI18}; failing to detect non-compositional phrases can hurt system effectiveness notably. Existing research treats phrases as either compositional or non-compositional in a deterministic manner. 
In this paper, we operationalize the viewpoint that compositionality is contextual rather than deterministic, i.e., that whether a phrase is compositional or non-compositional depends on its context. For example, the phrase ``green card'' is compositional when referring to a green colored card, whereas it is non-compositional when meaning permanent residence authorization.
We address the challenge of detecting this type of contextual compositionality as follows: given a multi-word phrase, we enrich the word embedding representing its semantics with evidence about its global context (terms it often collocates with) as well as its local context (narratives where that phrase is used, which we call \textit{usage scenarios}). We further extend this representation with information extracted from external knowledge bases. The resulting representation incorporates both localized context and more general usage of the phrase and allows to detect its compositionality in a non-deterministic and contextual way. Empirical evaluation of our model on a dataset of phrase compositionality\footnote{https://github.com/dswang2011/ImprovedRankedList/tree/master/input}, manually collected by crowdsourcing contextual compositionality assessments, shows that our model outperforms state-of-the-art baselines notably on detecting phrase compositionality.
\end{abstract}

%
%



\maketitle

\section{Introduction}
Automatic compositionality detection refers to the automatic assessment of the extent to which the meaning of a multi-word phrase is decomposable into the meanings of its constituents words and their combination. For example, while \textit{brown dog} is a fully compositional phrase meaning a dog of brown color, \textit{hot dog} is a non-compositional phrase denoting a type of food. Compositionality plays a vital role in word embeddings because a non-decomposable phrase should, in principle, be treated as a single word instead of a bag of word (BOW) in word embedding approaches. 

 A typical line of research in automatic compositionality detection is to "perturb" the input phrase by replacing one of its constituent words at a time with its synonym, and then to measure the semantic distance between the original phrase and the perturbed phrase set~\cite{Lioma2017}. The larger this distance, the less compositional the original phrase. For instance, \textit{hot dog} would be perturbed to \textit{warm dog} and \textit{hot canine}. The semantic distance between the original phrase and its two perturbations is high, indicating that they denote different concepts; hence \textit{hot dog} is non-compositional. However, the phrase \textit{brown dog} would be perturbed to \textit{hazel dog} and \textit{brown canine}, which have a shorter semantic distance to \textit{brown dog}, indicating that it is compositional.

In this paper, we posit that the compositionality of a phrase is not dichotomous or deterministic, but instead varies across scenarios. For instance, \textit{heavy metal} could refer to a dense metal that is toxic, which is compositional, but it could also be non-compositional when it refers to a genre of music. Previous work acknowledges this property of compositionality theoretically \cite{Lioma2017}, but no operational models implementing this have been presented to this day.

Given a multi-word phrase as input, we reason that the phrase is used in some narrative, e.g., a query, sentence, snippet, document, etc. We refer to this narrative as \textit{usage scenario} of the phrase. We combine evidence extracted from this usage scenario of the phrase with the global context (frequently co-occurring terms) of the phrase and use this to enrich the word embedding representation of the phrase. 
We linearly combine the weights of the tokens that are obtained from the usage scenario and the global context. We further extend this representation with information extracted from external knowledge bases.

We evaluate our model on a large dataset of phrases which are labeled as per five degrees of compositionality under various usage scenarios. We find that our model outperforms state-of-the-art baselines notably on identifying phrase compositionality. 
Our contributions are as follows:
\begin{itemize}
    \item  A novel model that detects phrase compositionality under different contexts and that outperforms the state of the art performance in the area.
    \item A benchmarking dataset of contextualized compositionality detection, that we make publicly available to the community.
\end{itemize}

\section{Related Work on Automatic Compositionality Detection} 
Compositionality detection mainly focuses on the semantic distance or similarity calculation between a given phrase and its component words or its perturbations under a corpus or dictionary. Earlier approaches mostly estimate the similarity between the original phrase and its component words. For example, Baldwin et al. \cite{baldwin2003empirical}, and Katz and Giesbrecht \cite{katz2006automatic} employ Latent Semantic Analysis (LSA) to calculate the semantic similarity (and hence to measure compositionality). Venkatapathy and Joshi \cite{venkatapathy2005measuring} extended this by adding collocation features, e.g., phrase frequency, point-wise mutual
information, extracted from the British National Corpus. 

More recent work estimates the similarity between a phrase and perturbed versions of that phrase where the words are replaced, one at a time, by their synonyms. For instance, Kiela and Clark \cite{Kiela13} compute the semantic distance between a phrase and its perturbation, using cosine similarity, which measures a phrase weight by pointwise-multiplication vectors of its terms. Lioma et al. \cite{lioma2015non} calculate the semantic distance with Kullback-Leibler divergence based on a language model; and, in subsequent work, Lioma et al. \cite{Lioma2017} represent the original phrase and its perturbations as ranked lists, and measure their correlation or distance. 

A promising line of work uses word embeddings and deep artificial neural networks for compositionality detection. Salehi \cite{salehi2015word} employs the word-based skip-gram model for learning non-compositional phrases, treating phrases as individual tokens with vectorial composition functions. Hashimoto and Tsuruoka \cite{hashimoto2016adaptive} adopt syntactic features including word index, frequency and PMI of a phrase and its components words to learn the embeddings. Yazdani et al. \cite{yazdani2015learning} utilize a polynomial projection function and deep artificial neural networks to learn the semantic composition and detect non-compositional phrases like those that stand out as outliers, assuming that the majority are compositional. 

Closer to our work, Salehi el. \cite{salehi2014detecting} use Wiktionary and utilize the definition, synonyms, and translations of Wiktionary to detect non-compositional components. Specifically, they analyze the lexical overlap between the definition of a phrase and its component words to measure compositionality. They assume that multi-word phrases are included in Wiktionary, while there is no guarantee for perfect coverage of the dictionary. Unlike this approach, we use Wiktionary together with DBPedia as a structured knowledge base to represent the contextual semantics of phrases.

To our knowledge, no prior work has operationalized the compositionality of a phrase as contextual.

\section{Our Contextual Representation Model for Compositionality Detection}
\label{sec:Model}


\begin{figure*}[h!]
\centering
\includegraphics[width=150mm]{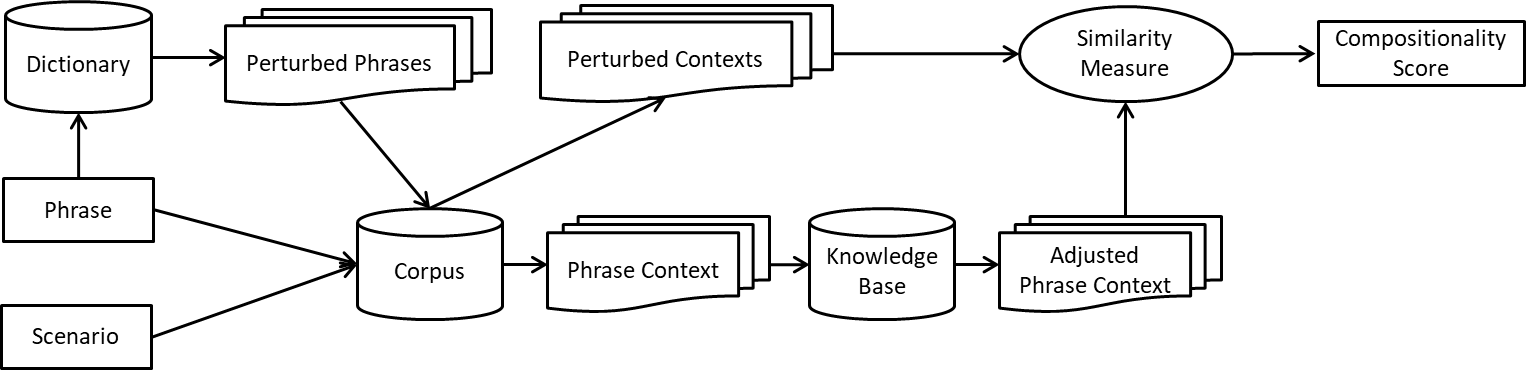}
\caption{The diagram for the CRM sequential framework.}
\label{fig:framework}
\end{figure*}

\subsection{Problem Formulation}
Given an input phrase $p$ and its accompanying usage scenario $s$, the aim is to compute the compositionality score $Score(p)$ of phrase $p$ with respect to usage scenario $s$.  We follow the substitution-based line of work~\cite{Kiela13}, which (a) generates perturbations of the input phrase $p$ by substituting one word at a time with its synonym, (b) builds a semantic representation (a vector of its co-occurring terms) separately for the input phrase $p$ and each perturbed phrase, and (c) uses the distance between the vectors of the input phrase and its perturbations to approximate the compositionality of the input phrase: the higher the distance, the less compositional the input phrase. This substitution-based line of work does not accommodate the usage scenario of the input phrase or its perturbations. The vectors of co-occurring terms are computed on one corpus, and hence these vectors represent the \textit{global} distributional semantics of the input phrase and its perturbations. We extend this line of work by incorporating the \textit{local} usage scenario of the phrase and its perturbations. We furthermore enrich these representations using external knowledge bases $KBs$. We describe this next.

Figure \ref{fig:framework} shows how the phrase and scenario are fed into the external corpus and knowledge base in a sequential manner in the architecture, which we refer to as contextual representation model (CRM).



\subsection{Building Global and Local Phrase Context}
\label{sec:local_context}
\paragraph{Global Phrase Context}
In Natural Language Processing (NLP), the distributional semantics of an input word are computed by fixing a natural number $n$ and, for each occurrence of a word in some corpus, finding the $n$ words occurring immediately before, and $n$ words occurring immediately after each occurrence of the input word (called \emph{context window}). If there is a total of $N$ context windows for a word, its distributional semantics in vector form can be calculated by using all these $N$ windows. Because this is a global representation of the word's distributional semantics across the whole corpus, the vector is called a \textit{globalized} vector. A general word embedding (e.g., word2vec) is comparable to such global context. 
Concretized representations of this globalized vector can be calculated with, e.g. ranked lists or word embeddings, as described in section \ref{ss:semantic_rep}. 

\paragraph{Local Phrase Context}
We aim to incorporate a representation of the local usage scenario of the input phrase (\textit{local phrase context}) into the above described global phrase context. This representation will not be in the vector directly, because usage scenarios are typically extremely short (in terms of words),  which may strongly bias the contextual representation of the phrase. Therefore, 
we rank all the global context windows of the phrase according to their similarity to the usage scenario of the phrase, and we
select the top $K$ most similar context windows to the usage scenario. These top $K$ context windows are used to build the local usage scenario context representation of the phrase. Then, we linearly combine the global representation of the phrase (i.e., by taking all $N$ context windows) and the local usage scenario representation of the phrase (i.e., the top $K$ context windows) to acquire the \textit{localized phrase context}. 
The ranking score is the similarity between the usage scenario $s$ and a window $W_i$, i.e. $\textrm{sim}_i= \textrm{similarity}(W_i,s) \in [0,1]$. The details of how the similarity score is computed are introduced in Section 4.

In the above, the value of $K$ is determined by the length of the usage scenario as follows:
\begin{equation}
       \label{eq:topN} K=\max\left(\frac{N}{2^{\textrm{length}(s)}}, M\right)
\end{equation}

\noindent where $N$ is the total number of context windows that contain phrase $p$ in the corpus; $\textrm{length}(s)$ is the number of words in usage scenario $s$ excluding the original phrase; and $M$ is a threshold. We explain these next. 

We posit that $2^{\textrm{length}(s)}$ indicates the degree of shrinking: the longer the usage scenario is, the smaller number of windows will be shrunk. The reason behind this is that the longer the usage scenario is, the more semantics it contains and subsequently fewer specific windows we are supposed to be capable of locating on. For instance, if the usage scenario is empty with $\textrm{length}(s)=0$, then it returns the entire $N$ windows of $p$ with no shrinking performed; if the usage scenario has three words with $2^{length(s)}=8$, it only collects the top $\frac{1}{8}$ of all the windows ($K=\frac{N}{8}$).
Note that $K$ depends on the usage scenario length, and is not fixed as a threshold of the similarity values. Since the similarity values may vary drastically in between $[0,1]$ for different usage scenarios, we argue that our method is more robust to such variations. Furthermore, an empirical threshold of $M$ is introduced to guarantee at least $M$ windows will be selected anyhow. To this end, the localized usage scenario context of a phrase is given by:
 \begin{eqnarray}
    \label{eq:local_context}
    \mathcal{C}(p,s)\ =
    \alpha\frac{\sum_{i=1}^{N} \mathcal{R}(W_i)}{N} + (1-\alpha)\frac{\sum_{j=1}^{K} \mathcal{R}(W_j)}{K}
\end{eqnarray}
where $a$ is a weight parameter between 0-1 indicating the weight of global vector, and the remaining of $1-a$ corresponds to the contribution of the localized vector; $\mathcal{R}$ denotes the semantic representation for $W$. In this paper, we represent a phrase as a ranked list of words and word embedding, so the same symbol $\mathcal{R}$ is adopted to denote both. The approach to calculate $\mathcal{R}$ is described in section \ref{ss:semantic_rep}.

\subsection{Enriching Global and Local Phrase Context with Knowledge Bases}
\label{sec:comb_kb}
We enrich the global and local context representations of the input phrase with information extracted from external knowledge bases. We describe this next. 

We reason that the corpus used to extract the global and local contexts has good coverage of various but not all possible usage scenarios of the phrase. A knowledge base is expected to contain more comprehensive, declarative information about the phrase, e.g., entities and phrase senses with categorized information. We, therefore, enrich the global and local phrase contexts extracted from the corpus with phrase information extracted from external knowledge bases. 

Given an input phrase, we collect all candidate senses and entities (uniformly referred to as \textit{candidates} in this paper) by searching  
the following properties (and associated values) from the knowledge bases: the properties \textit{dbpedia:redirects}, \textit{dbpedia:disambiguation} and their propagation relation with \textit{dbpedia:name} and \textit{rdfs:label}. The associated resources in these retrieved triples result in a set of candidates. Then, for each candidate, the values of \textit{rdfs:label}, \textit{dbpedia:abstract} and \textit{rdf:type} are concatenated as the context for that candidate, excluding the title (which is mostly the phrase name). We also use the interface\footnote{https://dkpro.github.io/dkpro-jwktl/} to retrieve senses from Wiktionary and merge them into the same candidate set for that given input.

Most phrases only contain a limited number of candidates, and different candidates of the same phrase can have entirely different meanings or be distinct entities. 
We hence investigate a sequential way to incorporate the knowledge base into the phrase contextual representation, as follows.
First, the phrase is fed into the knowledge base to find all candidate articles. Then, the candidates are ranked according to how similar they are to the localized phrase context. Those with similarity values above a certain threshold are identified as the matched candidates, denoted as $\{D_i, sim_i\}_{i=1}^n$, where $D_i$ and $\textrm{sim}_i$ refer to the $i^{th}$ matched candidate with similarity value $\textrm{sim}_i \in [0,1]$. A linear combination of the localized phrase context and the candidate articles is then conducted to compute the adjusted phrase context as follows:
    \begin{eqnarray}
        \label{eq:combi_seq}
        \mathcal{C}(p,D)\ = \lambda \mathcal{C}(p,s) + (1-\lambda) \sum_{i=1}^{n} w_i\mathcal{R}(D_i),
    \end{eqnarray}

\noindent where $w_i = \frac{\textrm{sim}_i}{\sum_{i=1}^n \textrm{sim}_i}$ is the normalized similarity score for the $i^{th}$ candidate article $D_i$ while $\mathcal{R}(D_i)$ denotes the semantic representation for $D_i$. Since KB contains well-defined knowledge of words, we use a weighted sum of the matched candidates, instead of a simple average of matched contexts in the text corpus.

The knowledge base we employ consists of DBpedia, Wiktionary, and Wordnet. DBpedia is constructed by extracting structured information from Wikipedia. The English version of the DBpedia contains 4.58 million entries, of which 4.22 million are classified and managed under one consistent ontology. Wiktionary is a multilingual, web-based, freely available dictionary, thesaurus and phrase book, designed as the lexical companion to Wikipedia. Volunteers collaboratively construct Wiktionary, so there are no specialized qualifications necessary. 

\subsection{Non-linear combination}

This section represents an approach of non-linear combination as a companion to the linear combination approach introduced in section \ref{sec:local_context} and \ref{sec:comb_kb}. In addition to the weight parameter oriented design of the linear combination, we also employ a non-linear sigmoid function in RNNs (recurrent neural networks), which resolves the arrangement of the combining order for context inputs. In other words, RNNs take into consideration the feedback from the previous context vector back and forth, leading to numerous applications \cite{HansenHASL19,HansenHASL19b}. Specifically, we train a neural network model using the Keras library to identify the compositionality label of each phrase. We encode the semantics adopting pre-trained word embeddings - word2vec \cite{word2vec} as word representations, a recurrent neural network with LSTM cells as the model, and cross-entropy as the loss function. As an optimizer, we utilize Adam optimizer for training the model. 

In a realistic scenario (also represented in our dataset) there are fewer non-compositional than compositional phrases. This situation resembles the class imbalance issue which happens when one class (or label) is represented by most of the examples while the other one is represented just by a few. Therefore, we adopt re-sampling strategies to tackle this problem.



\subsection{Compositionality Detection}
Here we introduce our proposed method for compositionality detection with Algorithm \ref{alg:A}.  Given a phrase $p$ of length $l$ and its usage scenario $s$ in a large corpus $Corp$, we compute its compositionality score through the following steps:  

\renewcommand{\algorithmicrequire}{ \textbf{Input:}} 
\renewcommand{\algorithmicensure}{ \textbf{Output:}} 


\begin{algorithm}[htb]
\caption{Algorithm of contextual compositionality detection }
\label{alg:A}
\begin{algorithmic}[1]
\REQUIRE Phrase $p$ with length $l$ \\
\REQUIRE Usage scenario $s$ for $p$ \\
\REQUIRE Corpus $Corp$ \\
\REQUIRE Knowledge Base - DBPedia  \\
\REQUIRE Similarity threshold - $thred$ \\
\ENSURE Compositionality score $comp(p)$ \\
\STATE {Set of perturbed phrase S($\hat{p}$)} $\leftarrow \varnothing$ \\
\STATE {Find synonym $\hat{t}$ of each term $t \in p$ }
\FOR{each $\hat{t}$}
\STATE Perturbed phrase $\hat{p} \leftarrow$ \{ $\hat{t}$ , $l\textrm{-1}$ original terms $t_o$ \}
\STATE Update perturbed phrase set $S(\hat{p}) \leftarrow  S(\hat{p}) \cup \hat{p}$
\ENDFOR
\FOR{phrase $p' \in \{ p \cup S(\hat{p})\}$}
\STATE $C(p') \leftarrow$ get context terms from localized phrase context from $Corp$, smooth with Eq. \ref{eq:topN} if it has scenario
\ENDFOR
\STATE Find $n$ candidate articles $D_i$ from KB where $sim_i=similarity(s,D_i)>thred$
\STATE $\mathcal{R}(D_i) \leftarrow$ semantic representation of $D_i$
\STATE $C(p,D) \leftarrow$ linear combined context $\lambda C(p) + (1-\lambda)\sum_{i=1}^{n} \frac{sim_i}{\sum_{i=1}^n sim_i} \mathcal{R}(D_i)$
\STATE $Q(L_{\hat{p}}) \leftarrow \varnothing$
\FOR{each perturbed phrase $\hat{p} \in S(\hat{p})$}
\STATE $Q(L_{\hat{p}})  \leftarrow Q(L_{\hat{p}}) \cup C(\hat{p})$
\ENDFOR
\RETURN $\frac{1}{|Q(L_{\hat{p}})|} \sum_{C(\hat{p}) \in Q(L_{\hat{p}})} Similarity(C(\hat{p}),C(p,D))$
\end{algorithmic}
\end{algorithm}

\begin{enumerate}
    \item Obtain localized phrase context through Eq.~\ref{eq:topN} and~\ref{eq:local_context}. The usage scenario of a phrase is the critical information, and the idea behind this step is to smooth the scenario context representation with the original phrase representation, as shown in line 8 in Algorithm \ref{alg:A}.
    
    
    \item Adjust phrase context with a knowledge base. The knowledge base is fed to adjust the localized phrase context where we adopt Eq.~\ref{eq:combi_seq} to encode the information (from line 10 to line 12).
    
    \item Obtain a perturbed phrase set. For each term in the phrase, we find its synonyms in WordNet. We then generate the set of perturbed phrases $S(p)$ as: S(p) = \{$\hat{p}$ where $\hat{p}$ = $l$-1 terms of p plus a synonym of the remaining term of p\}, from line 3 to line 6.
   
    \item Construct a perturbation representation set. For each perturbed phrase $\hat{p}$ in S(p), the corresponding representation $\mathcal{C}(\hat{p})$ is composed of all windows of $\hat{p}$ from the corpus, and is added to the perturbation list $Q(L_{\hat{p}})$ from line 14 to line 16. Note that we do not combine context from KB for perturbations. 
    
    
    \item Compute the compositionality score for the input phrase, shown in line 17, using the following equation:
    \begin{eqnarray}
        score(p) = \frac{\sum_{\hat{p} \in S(p)} sim(\mathcal{C}(p,D),\mathcal{C}(\hat{p}))}{|S(p)|}
    \end{eqnarray}

\end{enumerate}

\section{Implementation}
\label{sec:Implementation}
\subsection{Semantic Representation}
\label{ss:semantic_rep}
The semantic representation of a context (a phrase, a context window or a candidate content), i.e., $\mathcal{R}(\cdot)$ is concretized as either a ranked list or a word embedding. For the ranked list model, we calculate the TF-IDF as weight for all the tokens, rank them according to the weight, resulting in a ranked list of those tokens as the localized contextual representation. For the word embedding model, we use existing pre-trained word vectors - Glove \cite{pennington2014glove}, and represent the vector with the average of all tokens. The corpus we employ in our experiment is ClueWeb12-B13, a subset of some 50 million pages of ClueWeb12-Full dataset\footnote{https://lemurproject.org/clueweb12/}.

These two contextual representations lead to two different compositionality scores for the same model. We apply suffixes "-word embedding" or "-ranked list" in order to distinguish the way the contextual representation is computed, resulting in two distinct models, namely \textit{CRM word embedding} and \textit{CRM ranked list}.

\subsection{Similarity Measure}
In this study, we are faced with the problem of computing the similarity value between two context vectors. Here, we consider two types of similarity measures to achieve this purpose: \textit{cosine similarity} and \textit{Pearson correlation coefficient}.

One of the most commonly used similarity measures, \textit{cosine similarity}, computes the cosine value of the angle between the two vectors of the same length. For two vectors $\vec{a}=[a_1,a_2,..,a_n]$ and $\vec{b}=[b_1,b_2,..,b_n]$, their cosine similarity cossim(a,b) is given below:

\begin{equation}
    cossim(a,b) = \frac{\sum_{i=1}^n {a_ib_i}}{\sqrt{\sum_{i=1}^n a_i^2}\sqrt{\sum_{i=1}^n b_i^2}}
\end{equation}

The \textit{Pearson correlation coefficient} computes the degree of correlation between two variables, each having a set of observed values. Suppose two variables $X$ and $Y$ are associated with two set of values $\{X_1,X_2,...,X_n\}$ and $\{Y_1,Y_2,...,Y_n\}$ respectively. The Pearson correlation coefficient $r$ can, therefore, be computed as follows:
\begin{equation}
    r = \frac{\sum_{i=1}^n {(X_i-\overline{X})(Y_i-\overline{Y})} }{\sqrt{\sum_{i=1}^n {(X_i-\overline{X})^2}}\sqrt{\sum_{i=1}^n {(Y_i-\overline{Y})^2}}}
\end{equation}
where $\overline{X}$ and $\overline{Y}$ denote the average of $X$ and $Y$ respectively.

\subsection{Perturbation}

Here, we introduce the process to obtain the perturbations of a phrase $p$ with length $l$. First, we get the synonyms for each word in the phrase. Then, we construct the whole perturbation set, which contains all phrases composed of $l-1$ words in $p$ and a synonym of the remaining word. Suppose the $i^{th}$ word has $n_i$ synonyms, then the perturbation set contains $\sum_{i=1}^{l} n_i$ perturbed phrases. We then prune the perturbation set by filtering out the rare perturbed phrases in the text corpus. Basically, we compute the occurrence frequency of all perturbed phrases and pick the perturbed phrases with top $K$ frequency values. In our study, we set $K$ to be 7, which is derived from empirical observation of the data. Then, the final perturbation set contains 7 perturbed phrases in our study. 

\subsection{Parameter Settings}

\paragraph{Contextual Windows setting:} We set $window = 20$, which means it scans the previous 20 and subsequent 20 words of that phrase, with a sum of 40 words for each window. In Equation \ref{eq:topN}, we set $M=10$, and the base 2 can also be parameter-free which can be changed into 2,3,4,etc., to increase the localization level. 

\paragraph{Knowledge base threshold:} As for the threshold of KB candidates, we set the threshold of similarity value between localized context and KB candidates as 0.5 to filter out those candidates with similarity less than 0.5. 
\paragraph{Ranked list length:} In line with the work \cite{Lioma2017}, we set a maximum length of the ranked list as 1000, which means that we rank the tokens according to their TFIDF weight, and the tokens after the position of 1000 would be pruned. 

\begin{table*}[!h]
\minipage{0.49\textwidth}
\begin{tabular}{@{}llll@{}}
\toprule
\textbf{Unsupervised Methods}   & $\rho$ & $\alpha$, $\lambda$ \\ \midrule
Baseline: Ranked list \cite{Lioma2017} & 0.131&na \\
Baseline: Word Embedding & 0.147 & na \\
CRM ranked list & 0.209 & 0.1,0.5 \\
CRM Word Embedding     & 0.375    & 0.9, 0.1 \\
\toprule
\textbf{Supervised Methods} (20\% Testing) & $\rho$ & \\ \midrule
RNN (LSTM cells)      & 0.176    & na \\
RNN (LSTM cells) CRM      & 0.324    & na \\
\bottomrule
\end{tabular}
\caption{\label{tb:perform_compare}Results of different compositionality detection methods; \textit{na} denotes \textit{not applicable}.}
\endminipage\hfill
\minipage{0.49\textwidth}
\begin{tabular}{ll}
\hline
No. Non-Compositional & 43 (3.6\%)\\
No. Mostly Non-Compositional & 145 (12.1\%) \\
No. Ambiguos Phrases & 126 (10.5\%) \\
No. Mostly Compositional & 141 (12.0\%) \\
No. Compositional & 739 (61.8\%)\\
Unique number of Phrases & 1042 \\
No. of context & 1194 \\
Average number of context by Phrase & 1.146 \\ \hline
\end{tabular}
\caption{Summary of dataset statistics.}
\label{tab:data_stats}
\endminipage\hfill
\end{table*}

\paragraph{Training and testing:} As we are working with imbalanced data, we use a random oversampling strategy. We split our data in a stratified fashion into 65\% for training, 15\% for validation, and 20\% for testing. The re-sampling is be done after splitting the data into training and test, and only on the training data, i.e., none of the information in the test data is being used to create synthetic observations.

\section{Experiment and Validation}
In this section, we evaluate the effectiveness of the model presented in Section \ref{sec:Model}. Section 5.1 introduces the dataset and Section \ref{subsec:performance} presents the results achieved by our model.

\subsection{Crowdsourcing data}
We employ a dataset that consists of 1042 phrases that are noun-noun 2-term phrases \cite{farahmandSmithNivre2015data}. In this dataset, each phrase was assessed four times using a binary scale (compositional or non-compositional). However, these phrases are assessed with a deterministic label, meaning that no scenario or context was given, and the degree of compositionality may not always be binary \cite{Reddy:EtAl:11}. Therefore, we extend the dataset into a new version where each phrase is enriched with one or two scenarios if possible, by taking advantage of a crowdsourcing website - Figure Eight \footnote{https://www.figure-eight.com/}, and we use a graded level of compositionality. In Table ~\ref{tab:data_stats} we summarize the dataset statistics.

We divided the assessment into two stages: for the first stage, the trustful assessors, with level 3 (highest in Figure Eight), are required to understand the various meanings of a phrase, and, if possible, create two scenarios for the same phrase. From these two scenarios, one should be compositional or as compositional as possible, and the other non-compositional or as non-compositional as possible. If the phrase can only be compositional or only be non-compositional, then they create one scenario for it. For the second stage, the assessors are required to assess the compositionality of phrases within different scenarios with one of the five graded labels: compositional, mostly-compositional, ambiguous to judge, mostly non-compositional, and non-compositional. Note that, for the first stage, the two scenarios of a phrase are not necessarily of two extreme polarities. 


\begin{figure*}[!h]
\minipage{0.49\textwidth}
  \includegraphics[width=\linewidth]{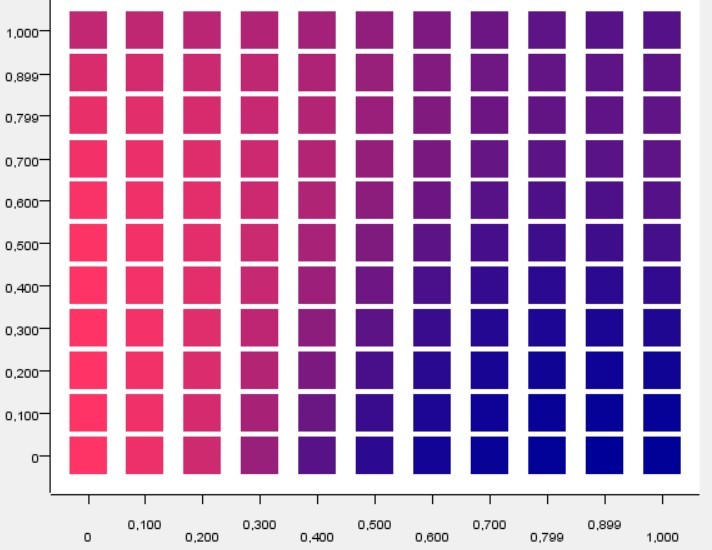}
  \caption{\label{fig:we_perfom} The grid search for Word embedding based Contextual representation. x-coordinate is $\alpha$ for controlling localized context and  $\lambda$ stands for y-coordinate, controlling the KB combining weight. Deeper blue represents higher performance whereas red indicates the opposite.}
\endminipage\hfill
\minipage{0.49\textwidth}%
  \includegraphics[width=\linewidth]{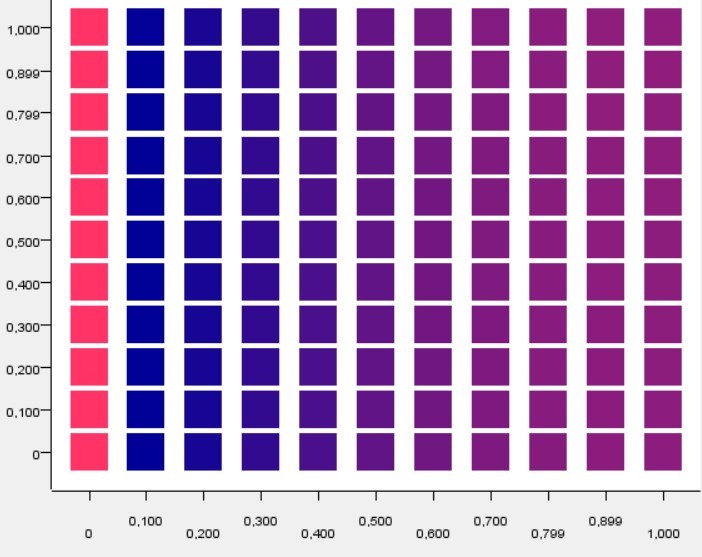}
  \caption{\label{fig:rl_perfom}The grid search for ranked list based Contextual representation. x-coordinate is $\alpha$ for controlling localized context and y-coordinate stands for $\lambda$, controlling the KB combining weight. Deeper blue represents higher performance whereas red indicates the opposite.  }
\endminipage
\end{figure*}
\subsection{Performance and Validation}
\label{subsec:performance}
Two linear combination parameters influence the performance of our model: the combination weight $\alpha$ (in Eq.~\ref{eq:local_context}) between the vectors of a phrase and its scenario, resulting in a localized phrase context, and $\lambda$ (in Eq.~\ref{eq:combi_seq}) between the localized context and knowledge base. The impacts of these two parameters on the final performance are visualized in Figure \ref{fig:we_perfom} and \ref{fig:rl_perfom}, corresponding to the word embedding-based and ranked list-based contextual representation respectively. $\alpha$ and $\lambda$ denote the $x$ and $y$ coordinates. The colors indicate the performance, which is the correlation between the ground truth labels and the predicted labels of our models ranging from -1 to 1. The performance values are colored ranging from red to blue, representing the lowest performance to the highest. 

As shown in Figure \ref{fig:we_perfom}, the performance is negatively correlated with $\alpha$ while positively correlated with $\lambda$. This indicates that reducing the relative importance of localized context (right direction on x-coordinate) while enhancing the influence of knowledge base (bottom direction on y-coordinate) can improve the performance for word embedding based contextual representation. In contrast, as shown in Figure \ref{fig:rl_perfom}, if we ignore the \textit{0} column, $\alpha$ is negatively correlated with the performance while $\lambda$ does not have an apparent influence on the performance. This indicates that attaching higher importance to the localized context (left direction on x-coordinate) can improve the performance for the ranked list based contextual representation, while the adoption of knowledge base does not have an apparent influence to the overall performance. The first \textit{0} column, which is shown more like an outlier, indicates that the existence of a vector of the original phrase is necessary. In other words, localized context would have relatively poor performance.  

As summarized in Table \ref{tb:perform_compare}, the performance improved from 0.147 to 0.375 for CRMs based on word embeddings (the best); from 0.131 to 0.209 for CRMs based on ranked list; and 0.176 to 0.324 for CRM based on RNN. For the word embedding based contextual representation model, relying more on the knowledge base while keeping the scenario to limited importance will lead to a high-performed model; for the ranked list based contextual representation model, on the other hand, adequately high adoption of localized context can lead to improved performance. The reason behind this can be that the knowledge base contains relatively trimmed but well-categorized information, therefore, the word embedding model can take full use of this text as informative vectors. In contrast, ranked lists, depending on tokens, work better on a large-scale corpus where they induce a large number of context windows. However, the knowledge base contains a limited number of tokens that may have little contribution to the final representation. Even though we can tune the weight of tokens from a knowledge base, it still can have limited influence in comparison to the long ranked list, which can be as long as 1000 tokens in our experiment.

For the non-linear combination where we employed the sigmoid function in RNNs, the CRM based on RNN still beats the original RNN. However, the performance is still lower than the unsupervised approaches. 



\section{Conclusion}
We developed a novel method for compositionality detection where the compositionality of a phrase is contextual rather than static. Instead of considering an isolated phrase as input, we assume a phrase and its usage scenario (e.g., a query, snippet, sentence, etc.) as input, and we model a joint semantic representation of these by combining distributional semantics extracted from a corpus and additional evidence extracted from an external structured knowledge base. 

Our resulting model uses word embeddings to detect compositionality, more accurately than the related state of the art. Our experiments show that for word embeddings, the usage of knowledge bases can lead to notable performance improvements. 

In the future, we plan to evaluate our model on further datasets and compositionality detection scenario, e.g., Verbal Phraseological Units (VPUs).

\section*{ACKNOWLEDGEMENT}
This work is supported by the Quantum Access and Retrieval Theory (QUARTZ) project, which has received funding from the European Union's Horizon 2020 research and innovation programme under the Marie Sklodowska-Curie grant agreement No. 721321.

\bibliographystyle{ACM-Reference-Format}
\balance 
\bibliography{sample-base}

\end{document}